# Theedhum Nandrum@Dravidian-CodeMix-FIRE2020: A Sentiment Polarity Classifier for YouTube Comments with Code-switching between Tamil, Malayalam and English

BalaSundaraRaman Lakshmanan[a], Sanjeeth Kumar Ravindranath[b]

[a]*DataWeave, Bengaluru, India*
[b]*Exotel, Bengaluru, India*

**Abstract**
Theedhum Nandrum is a sentiment polarity detection system using two approaches–a Stochastic Gradient Descent (SGD) based classifier and a Long Short-term Memory (LSTM) based Classifier. Our approach utilises language features like use of emoji, choice of scripts and code mixing which appeared quite marked in the datasets specified for the Dravidian Codemix - FIRE 2020 task. The hyperparameters for the SGD were tuned using GridSearchCV. Our system was ranked 4[th] in Tamil-English with a weighted average F1 score of 0.62 and 9[th] in Malayalam-English with a score of 0.65. We achieved a weighted average F1 score of 0.77 for Tamil-English using a Logistic Regression based model after the task deadline. This performance betters the top ranked classifier on this dataset by a wide margin. Our use of language-specific Soundex to harmonise the spelling variants in code-mixed data appears to be a novel application of Soundex. Our complete code is published in github at https://github.com/oligoglot/theedhum-nandrum.

## 1. Introduction

Dravidian languages are spoken by 227 million people in south India and elsewhere. To improve production of and access to information for user-generated content of Dravidian languages [1, 2] organised a shared task. *Theedhum Nandrum* [1] was developed in response to the Dravidian-CodeMix sentiment classification task collocated with FIRE 2020. We were supplied with manually labelled training data from the datasets described in TamilMixSentiment [3] and MalayalamMixSentiment [4]. The datasets consisted of 11,335 training, 1,260 validation and 3,149 test records for Tamil-English code-mixed data and 4,851 training, 541 validation and 1,348 test records for Malayalam-English code-mixed data.

The comments in the dataset exhibited inter-sentential switching, intra-sentential switching and tag switching [5, 6]. Even though Tamil and Malayalam have their own native scripts [7], most comments were written in Roman script due to ease of access to English Keyboard [8]. The comments often mixed Tamil or Malayalam lexicons with an English-like syntax or vice versa. Some comments were written in native scripts but with intervening English expressions. Even though these languages are spoken by millions of people, they are still under-resourced and there are not many data sets available for code-mixed Dravidian languages [9, 10, 11].

---

[1]*Theedum Nandrum* is a phrase from the 1st century BCE Tamil literary work Puṟanāṉūṟu. Meaning "the good and the bad", it is part of the oft-quoted lines *"All towns are ours. Everyone is our kin. Evil and goodness do not come to us from others."* written by Kaṇiyan Pūngunṟanār.

## 2. Method

Given the particular register of language used in YouTube comments and the fact that most of the comments used the Roman alphabet to write Tamil and Malayalam text without following any canonical spelling, we understood the importance of pre-processing and choice of features over other specifics of the Machine Learning model to be used. This was evident from the bench marking results on the gold dataset in TamilMixSentiment [3]. We used core libraries like Keras [2] and scikit-learn [3] for the classifiers.

### 2.1. Pre-processing

We normalised the text using The Indic Library [4] to canonicalise multiple ways of writing the same phoneme in Unicode. We also attempted spelling normalisation by doing a brute force transliteration from Roman to Tamil or Malayalam, followed by a dictionary lookup using a SymSpell-based spell checker on a large corpus [5]. However, we did not get much success in finding dictionary matches up to edit distance 2, the highest supported value. We then chose to use an Indian language specific Soundex as a feature to harmonise the various spellings with some success as described in 2.2.2.

Words from multiple corpora indexed by their Soundex values could be used to get canonical spellings where there is long-range variation. We can combine edit distance allowance and Soundex equivalence while looking up our dictionary. The potential utility of such a method is supported by the characterisation of the text of these datasets in [12].

### 2.2. Feature Generation

#### 2.2.1. Emoji Sentiment

We noticed that a key predictor of the overall sentiment of a comment was the set of emoji used. Based on this observation, we extracted the emoji from text and used Sentimoji [13] to assign a sentiment (positive, negative or neutral) to the emoji. However, the list of emoji available in Sentimoji did not include a majority of the emoji found in our datasets. We used the sentiment labels in the training data to compute a sentiment polarity for each of the missing emoji based on the frequency of use in each class. We used both the raw emoji as well as its inferred sentiment as features.

#### 2.2.2. Soundex

As mentioned previously in 2.1, we used Soundex to harmonise the numerous spelling variants of the same word when expressed in the Roman alphabet. For example, the Tamil word நன்றி is written as *nanri* and *nandri* in the corpus. The standard Soundex algorithm for English did not approximate Tamil and Malayalam words well. We found libindic-soundex [6] to perform very well. Soundex has been employed in spoken document classification [14, 15] where it helps in learning over transcription errors. Our use of language-specific Soundex to harmonise the spelling variants in code-mixed data appears to be a first of its kind.

---

[2] https://github.com/fchollet/keras
[3] https://github.com/fchollet/keras
[4] https://github.com/anoopkunchukuttan/indic_nlp_library/blob/master/docs/indicnlp.pdf
[5] https://github.com/indicnlp/solthiruthi-sothanaikal
[6] https://github.com/libindic/soundex

The specificity improves when the input text was in Tamil or Malayalam script rather than in Roman alphabets. Hence, we used indictrans [16] to perform a transliteration to native scripts before feeding the text to the Soundex generator function. That gave improved matches. For example, அருமை and அரும have a Soundex of அ PCND000, while *arumai* in Roman alphabets gets a65. This problem is mitigated by using indictrans as above before generating the Soundex values.

### 2.2.3. Language Tag

Comments were not all in the expected language of the dataset. Some were in other languages either using their native scripts or the Roman alphabet. The classifier was expected to label `not Tamil` or `not Malayalam` as the case may be. To support that as well as to ensure the features specific to a language are aligned well, predicted language from Google Translation API [7] was added as a feature. Tagging parts of the code-mixed comments into respective languages should improve the classification accuracy further.

### 2.2.4. Word Vector

We tokenised the text based on separators, but retained most other characters so as to not drop any non-word signals. We also added word ngrams up to length 4 as features.

### 2.2.5. Document length range

We bucketed Document length into 21 ranges viz. `1-10, 11-20,...,>200` was used as a feature. This improved the performance.

## 2.3. Classifiers

The task required us to classify the comments into 5 classes viz. `mixed_feelings, negative, positive, not-tamil/not-malayalam, unknown_state`. After evaluating various other linear models, we picked SGD as the best performing algorithm for the data at hand with the features we had used at the time of benchmarking. Additionally, we trained an LSTM-based classifier [17] which did not perform as well as the linear classifier. A combined approach may perform better in the face of text mixed with multi-modal noise [18].

### 2.3.1. Stochastic Gradient Descent (SGD) Classifier

Based on parameter tuning, we arrived at the following configuration which gave the best performance on trials using the training dataset. An SGD classifier [19] with modified Huber loss and a learning rate of `0.0001` was used. Different weights were applied to the features of Tamil and Malayalam.

---

[7]https://cloud.google.com/translate/docs/reference/rest/v3/projects/detectLanguage

**Table 1**
*Theedhum Nandrum* Performance. W-weighted average

| Language | Dataset | W-Precision | W-Recall | W-F1 Score |
|---|---|---|---|---|
| Tamil | Validation (SGD) | 0.74 | 0.65 | 0.68 |
| | Validation (LSTM) | 0.46 | 0.68 | 0.55 |
| | Test | 0.64 | 0.67 | 0.62 |
| Malayalam | Validation (SGD) | 0.73 | 0.64 | 0.67 |
| | Validation (LSTM) | 0.17 | 0.41 | 0.24 |
| | Test | 0.67 | 0.66 | 0.65 |

### 2.3.2. Long Short-term Memory (LSTM)

A 4-layer sequential model was trained. Embedding, SpatialDropout, LSTM and a Densely-connected Neural Network were the layers. Softmax was used in the last layer to generate probability distribution on all classes. We used categorical cross entropy loss and Adam optimiser with a learning rate of 0.0001. The learning seemed to maximise at 15 epochs for Tamil and 10 for Malayalam. Based on the results in Table 1, we found that it performed worse than the SGD Classifier. We identified that there was considerable overfitting because of the class imbalance in the relatively small training dataset. A pre-trained embedding combined with transfer learning could improve the performance [20].

### 2.4. Parameter Tuning

Tuning and optimisation of the SGD model was performed using grid-based hyper-parameter tuning. Since a `FeatureUnion` of Transformers was used with a Stochastic Gradient Classifier, two types of parameters were optimised.

1. Parameters of the Classifier
2. Weights of the Transformers in the FeatureUnion

For the Classifier, the main parameters that were optimised are the loss function and regularisation term (penalty). Tuning was also performed on the weights of the transformers of the FeatureUnion. The features used by the model are mentioned in 2.2. We observed that though the features used for classification were common to both Tamil and Malayalam language documents, the classification accuracy improved with different weights for the features for Tamil and Malayalam. For e.g., having a higher weight for Document Length Range (mentioned in 2.2.5) improved results for Malayalam.

## 3. Results

We tuned our SGD and LSTM classifiers using the available training data against the validation sets. We then classified the unlabelled test data using the optimised classifiers. We submitted the output from three of our best performing configurations between LSTM and SGD classifiers. The results for the test data were from the task organisers who picked the best of 3 classifications. The combined results are tabulated in Table 1.

The above results are better than the benchmark done in TamilMixSentiment [4]. *Theedhum Nandrum* was ranked 4[th] in the Dravidian-CodeMix task competition for Tamil-English, the weighted average F1 score was only 0.03 less than the top ranked team SRJ [8]. With an average F1 score of 0.65, *Theedhum Nandrum* was ranked 9[th] in the Dravidian-CodeMix task competition for Malayalam-English [9].

---
[8]https://dravidian-codemix.github.io/2020/Dravidian-Codemix-Tamil.pdf

**Table 2**
*Theedhum Nandrum* Logistic Regression Model Performance. W-weighted average

| Language | Dataset | W-Precision | W-Recall | W-F1 Score |
|---|---|---|---|---|
| Tamil | Validation | 0.91 | 0.70 | 0.78 |
|  | Test | 0.91 | 0.68 | 0.77 |
| Malayalam | Validation | 0.80 | 0.54 | 0.61 |
|  | Test | 0.73 | 0.67 | 0.69 |

After the task deadline, we ran a benchmark on other linear models with the full set of features above. Logistic Regression performed much better giving a weighted average of 0.77 for Tamil and 0.69 for Malayalam with the following parameters `C=0.01, penalty='l2', solver='newton-cg'` as shown in Table 2. Since we picked SGD based on our benchmarking performed before we added the Soundex feature, we had overlooked this better performing configuration. The performance of this classifier even exceeds the top ranked-classifier for the Tamil-English dataset by a wide margin.

## 4. Conclusion

*Theedhum Nandrum* demonstrates that SGD and Logistic Regression based models leveraging spelling harmonisation achieved by using language-specific Soundex values as features for code-mixed text perform well on the code-mixed datasets specified for the Dravidian Codemix - FIRE 2020 task. Our use of language-specific Soundex to harmonise the spelling variants in code-mixed data appears to be a first of its kind. In addition, emoji are a useful feature in sentiment prediction over YouTube comments. Future work is required to validate the usefulness of spelling correction using a combination of edit distance and Soundex.

## Acknowledgments

The authors are grateful for the contributions of Ishwar Sridharan to the code base. Particularly, the idea of using emoji as a feature is owed to him, among other things. We also thank Shwet Kamal Mishra for his inputs relating to LSTM. The logo for *Theedhum Nandrum* software was designed by Tharique Azeez signifying the duality of good and bad using the yin yang metaphor. *Theedhum Nandrum* also benefited from open source contributions of several people.

## References

[1] B. R. Chakravarthi, R. Priyadharshini, V. Muralidaran, S. Suryawanshi, N. Jose, J. P. Sherly, Elizabeth McCrae, Overview of the track on Sentiment Analysis for Dravidian Languages in Code-Mixed Text, in: Working Notes of the Forum for Information Retrieval Evaluation (FIRE 2020). CEUR Workshop Proceedings. In: CEUR-WS. org, Hyderabad, India, 2020.

---

[9]https://dravidian-codemix.github.io/2020/Dravidian-Codemix-Malayalam.pdf

[2] B. R. Chakravarthi, R. Priyadharshini, V. Muralidaran, S. Suryawanshi, N. Jose, J. P. Sherly, Elizabeth McCrae, Overview of the track on Sentiment Analysis for Dravidian Languages in Code-Mixed Text, in: Proceedings of the 12th Forum for Information Retrieval Evaluation, FIRE '20, 2020.

[3] B. R. Chakravarthi, V. Muralidaran, R. Priyadharshini, J. P. McCrae, Corpus creation for sentiment analysis in code-mixed Tamil-English text, in: Proceedings of the 1st Joint Workshop on Spoken Language Technologies for Under-resourced languages (SLTU) and Collaboration and Computing for Under-Resourced Languages (CCURL), European Language Resources association, Marseille, France, 2020, pp. 202–210. URL: https://www.aclweb.org/anthology/2020.sltu-1.28.

[4] B. R. Chakravarthi, N. Jose, S. Suryawanshi, E. Sherly, J. P. McCrae, A sentiment analysis dataset for code-mixed Malayalam-English, in: Proceedings of the 1st Joint Workshop on Spoken Language Technologies for Under-resourced languages (SLTU) and Collaboration and Computing for Under-Resourced Languages (CCURL), European Language Resources association, Marseille, France, 2020, pp. 177–184. URL: https://www.aclweb.org/anthology/2020.sltu-1.25.

[5] P. Ranjan, B. Raja, R. Priyadharshini, R. C. Balabantaray, A comparative study on code-mixed data of indian social media vs formal text, in: 2016 2nd International Conference on Contemporary Computing and Informatics (IC3I), 2016, pp. 608–611.

[6] B. R. Chakravarthi, M. Arcan, J. P. McCrae, Improving wordnets for under-resourced languages using machine translation, in: Proceedings of the 9th Global WordNet Conference (GWC 2018), 2018, p. 78.

[7] B. R. Chakravarthi, M. Arcan, J. P. McCrae, Wordnet gloss translation for under-resourced languages using multilingual neural machine translation, in: Proceedings of the Second Workshop on Multilingualism at the Intersection of Knowledge Bases and Machine Translation, 2019, pp. 1–7.

[8] B. R. Chakravarthi, M. Arcan, J. P. McCrae, Comparison of different orthographies for machine translation of under-resourced Dravidian languages, in: 2nd Conference on Language, Data and Knowledge (LDK 2019), Schloss Dagstuhl-Leibniz-Zentrum fuer Informatik, 2019.

[9] N. Jose, B. R. Chakravarthi, S. Suryawanshi, E. Sherly, J. P. McCrae, A survey of current datasets for code-switching research, in: 2020 6th International Conference on Advanced Computing and Communication Systems (ICACCS), 2020, pp. 136–141.

[10] R. Priyadharshini, B. R. Chakravarthi, M. Vegupatti, J. P. McCrae, Named entity recognition for code-mixed indian corpus using meta embedding, in: 2020 6th International Conference on Advanced Computing and Communication Systems (ICACCS), 2020, pp. 68–72.

[11] B. R. Chakravarthi, P. Rani, M. Arcan, J. P. McCrae, A survey of orthographic information in machine translation, arXiv e-prints (2020) arXiv–2008.

[12] B. R. Chakravarthi, Leveraging orthographic information to improve machine translation of under-resourced languages, Ph.D. thesis, NUI Galway, 2020. URL: http://hdl.handle.net/10379/16100.

[13] P. K. Novak, J. Smailović, B. Sluban, I. Mozetič, Sentiment of emojis, PLOS ONE 10 (2015) e0144296. URL: https://doi.org/10.1371/journal.pone.0144296. doi:10.1371/journal.pone.0144296.

[14] P. Dai, U. Iurgel, G. Rigoll, A novel feature combination approach for spoken document classification with support vector machines, in: Proc. Multimedia information retrieval workshop, Citeseer, 2003, pp. 1–5.

[15] M. A. Reyes-Barragán, L. Villaseñor-Pineda, M. Montes-y Gómez, A soundex-based approach for spoken document retrieval, in: Mexican International Conference on Artificial Intelligence,

Springer, 2008, pp. 204–211.
[16] I. A. Bhat, V. Mujadia, A. Tammewar, R. A. Bhat, M. Shrivastava, IIIT-H System Submission for FIRE2014 Shared Task on Transliterated Search, in: Proceedings of the Forum for Information Retrieval Evaluation, FIRE '14, ACM, New York, NY, USA, 2015, pp. 48–53. URL: http://doi.acm.org/10.1145/2824864.2824872. doi:10.1145/2824864.2824872.
[17] S. Hochreiter, J. Schmidhuber, Long short-term memory, Neural Computation 9 (1997) 1735–1780. URL: https://doi.org/10.1162/neco.1997.9.8.1735. doi:10.1162/neco.1997.9.8.1735.
[18] P. Agrawal, A. Suri, NELEC at SemEval-2019 task 3: Think twice before going deep, in: Proceedings of the 13th International Workshop on Semantic Evaluation, Association for Computational Linguistics, Minneapolis, Minnesota, USA, 2019, pp. 266–271. URL: https://www.aclweb.org/anthology/S19-2045. doi:10.18653/v1/S19-2045.
[19] T. Zhang, Solving large scale linear prediction problems using stochastic gradient descent algorithms, in: Twenty-first international conference on Machine learning - ICML '04, ACM Press, 2004. URL: https://doi.org/10.1145/1015330.1015332. doi:10.1145/1015330.1015332.
[20] P. Liu, W. Li, L. Zou, NULI at SemEval-2019 task 6: Transfer learning for offensive language detection using bidirectional transformers, in: Proceedings of the 13th International Workshop on Semantic Evaluation, Association for Computational Linguistics, Minneapolis, Minnesota, USA, 2019, pp. 87–91. URL: https://www.aclweb.org/anthology/S19-2011. doi:10.18653/v1/S19-2011.